\pdfoutput=1

\documentclass[11pt]{article}

\usepackage{acl}

\usepackage{dialogue}

\usepackage{times}
\usepackage{latexsym}
\usepackage{amssymb}
\usepackage[T1]{fontenc}
\usepackage[utf8]{inputenc}

\usepackage{microtype}
\usepackage[compact]{titlesec}
\usepackage{graphicx}
\usepackage{tabularx}
\usepackage{booktabs}
\usepackage{siunitx}
\DeclareSIUnit[number-unit-product = ]\percent{\char`\%}

\usepackage{etoolbox}
\robustify\bfseries

\usepackage[acronym]{glossaries}
\newacronym[plural={CEOs},firstplural=chief executive officers (CEOs)]{ceo}{CEO}{chief executive officer}
\newacronym[plural={CFOs},firstplural=chief financial officers (CFOs)]{cfo}{CFO}{chief financial officer}
\newacronym{hr}{HR}{human resources}

\newacronym[plural={ECs},firstplural=earnings calls (ECs)]{ec}{EC}{earnings call}
\newacronym{qa}{Q\&A}{questions-and-answers}

\newacronym{btm}{BTM}{Book-to-Market}
\newacronym{roa}{ROA}{Return on Assets}
\newacronym{sue}{SUE}{Standardized Unexpected Earnings}
\newacronym{vix}{VIX}{CBOE Volatility Index}

\newacronym{hci}{HCI}{human--computer interaction}
\newacronym{nlp}{NLP}{natural language processing}

\newacronym{hitl}{HITL}{human-in-the-loop}
\newacronym{liwc}{LIWC}{Linguistic Inquiry and Word Counts}

\newacronym{iq}{IQ}{intelligence quotient}
\newacronym[plural={MBTI},firstplural=Myers--Briggs Type Indicators (MBTI)]{mbti}{MBTI}{Myers--Briggs Type Indicator}

\newacronym{aic}{AIC}{Akaike information criterion}
\newacronym{bic}{BIC}{Bayesian information criterion}
\newacronym{iaa}{IAA}{Inter-Annotator Agreement}
\newacronym{mae}{MAE}{mean absolute error}
\newacronym{mse}{MSE}{mean-squared error}
\newacronym{rmse}{RMSE}{root-mean-square-error}

\newacronym{rbf}{RBF}{radial basis function}
\newacronym[plural={MLPs},firstplural=multilayer perceptrons (MLPs)]{mlp}{MLP}{multilayer perceptron}
\newacronym[plural={SVMs},firstplural=support vector machines (SVMs)]{svm}{SVM}{support vector machine}
\newacronym{tfidf}{tf--idf}{term frequency--inverse document frequency}
\glsdisablehyper

\usepackage{textcomp}  
\usepackage{scalerel}  
\def\huggingface{\kern1pt\scalerel*{\includegraphics{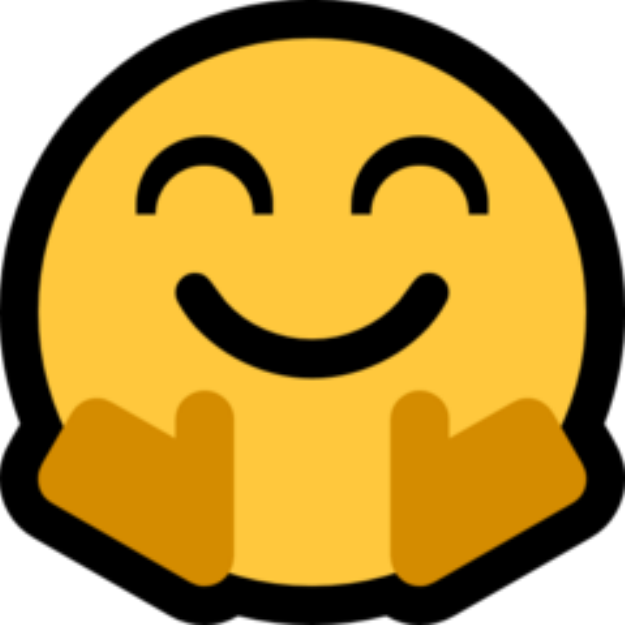}}{\textrm{\textbigcircle}}\kern3pt}

\usepackage{stfloats}
\usepackage{float}
\usepackage{subcaption}
\usepackage{multirow}
\usepackage{adjustbox}

\usepackage[symbol]{footmisc}

\title{Top-Down Influence? Predicting CEO Personality and \\ Risk Impact from Speech Transcripts}

\author{
Kilian Theil\textmd{\textsuperscript{\dag*}, }Dirk Hovy\textmd{\textsuperscript{\ddag}, }Heiner Stuckenschmidt\textmd{\textsuperscript{\dag}}\\ 
\textsuperscript{\dag}Data and Web Science Group, University of Mannheim, Germany\\
\textsuperscript{\ddag}MilaNLP, Bocconi University, Milan, Italy\\
\{kilian, heiner\}@informatik.uni-mannheim.de \\ 
dirk.hovy@unibocconi.it
}

\begin{document}
\maketitle

\begin{abstract}
How much does a CEO's personality impact the performance of their company? Management theory posits a great influence, but it is difficult to show empirically\textemdash there is a lack of publicly available self-reported personality data of top managers. Instead, we propose a text-based personality regressor using crowd-sourced \gls{mbti} assessments. The ratings have a high internal and external validity and can be predicted with moderate to strong correlations for three out of four dimensions. Providing evidence for the \textit{upper echelons theory}, we demonstrate that the predicted CEO personalities have explanatory power of financial risk.
\end{abstract}
\glsreset{mbti}

\footnotetext[1]{A part of this work was conducted during a research visit to MilaNLP Lab at Bocconi University in Milan.}
\renewcommand\thefootnote{\arabic{footnote}}

\section{Introduction}
How much influence does the personality of a \gls{ceo} have on their company's performance? The personal news and antics of famous \glspl{ceo} like Elon Musk, Jeff Bezos, or Bill Gates make headlines, and their personalities sometimes generate a cult-like following. But what measurable effect do they really have? The \textit{upper echelons theory} \cite{Hambrick1984} suggests that the personalities of \glspl{ceo} also reflect in the organizational outcomes of their companies. However, presumably due to the lack of labeled data, no supervised models exist to detect \glspl{ceo}' personalities from text and infer their effect on the financial performance of companies. In this paper, we close this research gap by presenting the first Transformer-based model to predict the impact of \glspl{ceo}' \gls{mbti} personality on financial risk.

Ideally, personality is assessed with self-reported questionnaires. However, it is technically infeasible to request executives such as Elon Musk to fill out targeted pen and paper questionnaires. We were therefore motivated to explore crowd-sourced data. This approach is supported by past research showing that observer reports are an inexpensive and valid alternative to self-reports \citep{VAZIRE2006472}, as they usually agree with them \citep{doi:10.1177/0956797618810000}, and are particularly suitable for the assessment of top management personality \citep{https://doi.org/10.1111/j.1468-2389.2007.00371.x}. 

The dominant personality model is the Big 5, which presents personality on a continuum along the dimensions \textit{openness}, \textit{conscientiousness}, \textit{extraversion}, \textit{agreeableness}, and \textit{neuroticism} \cite{McCrae1992}. The available data source we use lacks Big 5 ratings, so as proxy, we explore the \gls{mbti} \citep{Briggs-Myers1995}, which has been shown to correlate along the main dimensions with the Big 5 \citep{McCrae1989, Furnham1996, Furnham2003}. This model represents personality via the categories \textit{extraversion--introversion}, \textit{sensing--intuition}, \textit{thinking--feeling}, and \textit{judging--perceiving}. Addressing methodological criticism of the \gls{mbti} \cite{McCrae1989}, we 
\begin{itemize}
    \item explore an alternative \gls{mbti} representation as a vector of continuous values (\S\ref{data:pers});
    \item find a high internal and external validity of this measure (\S\ref{met:iaa});
    \item show that it can be predicted from text (\S\ref{res:pers_pred});
    \item and demonstrate that it is predictive of financial risk (\S\ref{res:risk}).
\end{itemize}

Overall, our findings lend empirical support to the \textit{upper echelons theory} of management.

\section{Background and Related Work}
Various personality measures exist in the literature. This section describes the personality model we explore (\gls{mbti}), the de-facto standard model (Big 5), and approaches to predict both representations of personality from text.

\subsection{MBTI}
The \gls{mbti} is named after Katherine Cook Briggs and Isabel Briggs Myers. They developed it based on the work of the analytical psychologist Carl Jung \cite{Briggs-Myers1995}. The \gls{mbti} classifies personalities binarily along the following axes:
\begin{itemize}
    \item \textit{extraversion} vs.\ \textit{introversion} (E--I): describing an out- or inward-oriented social attention;
    \item \textit{sensing} vs.\ \textit{intuition} (S--N): information processing based on perceivable/known facts or conceptualization and imagination;
    \item \textit{thinking} vs.\ \textit{feeling} (T--F): decision-making based on logic and rationality or emotions and empathy;
    \item \textit{judging} vs.\ \textit{perceiving} (J--P): quick judgement and organized action or observation and improvisation on-the-go.
\end{itemize}
Combined, the four labels form one of 16  personality types (e.g., ``ENTJ''). The \gls{mbti} is widely used in human resources management and by laypeople as a tool for self-exploration.

Psychological literature, however, has called assumptions of the \gls{mbti} into question. For example, \newcite{McCrae1989} find no evidence that personality can be binarized or distinguished into 16 different types. In addition, they find moderate to strong correlations between MTBI and Big 5 \cite{Mccrae2010}, which is described in greater detail below (\S\ref{big5}). We re-assess these correlations in our dataset and explore a continuous representation of the \gls{mbti} in line with the Big 5.

\paragraph{MBTI Prediction from Text}
In a literature study on text-based personality detection and a subsequent annotation study, \newcite{Stajner2020, Stajner2021} conclude that predicting the \gls{mbti} from textual data is a difficult task. They hypothesize that this is due to the theoretical and qualitative origin of the index, which distinguishes it from the empirical and quantitative Big 5. In particular, the dimensions \textit{sensing} vs.\ \textit{intuition} (S--N) and \textit{judging} vs.\ \textit{perceiving} (J--P) depend on behavioral rather than linguistic signals \citep[p. 6291]{Stajner2020}.

In a field survey of project managers, \newcite{Cohen2013} show that managers are significantly more often of the \textit{intuitive} (N) and \textit{thinking} (T) type than the general population. We observe a similar pattern in our dataset (\S\ref{data:pers}, Figure \ref{fig:label_dist}). Classifying the \gls{mbti} of Twitter users based on count-based features, gender, and tweet $n$-grams, \newcite{Plank2015} outperform a majority class baseline for the E--I and the T--F dimensions. \newcite{Gjurkovic2018} predict the self-reported \gls{mbti} of Redditors with \gls{svm} and \gls{mlp} models based on linguistic and activity-level features. Their model outperforms a majority class baseline across all dimensions with the best results for E--I, followed by S--N, J--P, and T--F.

We compare the best-performing approaches identified by prior \gls{mbti} prediction studies ($n$-grams and \gls{liwc} dictionaries with \glspl{svm} and \glspl{mlp}) to Transformer architectures. Furthermore, we consider a different domain (spoken financial disclosures) and perform a regression instead of a classification.

\subsection{Big 5}
\label{big5}
The Big 5 are the established psychometric model. Here, personality is represented as a continuum along the five axes \textit{openness}, \textit{conscientiousness}, \textit{extraversion}, \textit{agreeableness}, and \textit{neuroticism} \cite{McCrae1992}.

\paragraph{Big 5 Prediction from Text}
As part of the \textit{myPersonality} project, \newcite{Kosinski2015} find that liked Facebook pages predict Big 5, IQ, and other personal characteristics to varying degrees. \newcite{Mairesse2007} create a text-based Big 5 prediction tool based on student essays and speech recordings.

\newcite{https://doi.org/10.1002/smj.2974} show that  CEOs' Big 5 personalities moderate the relationship between CEO compensation and risk-taking. \newcite{Hrazdil2020} use \textsc{IBM Watson Personality Insight} to predict the Big 5 of C-level executives in earnings calls and find that an executive's personality is associated with their risk tolerance and company audit fees. \newcite{doi:10.5465/amj.2018.0626} find that CEO Big 5 are related to perceived firm risk and shareholder value. Another finding is that CEO \textit{conscientiousness} moderates the effect of financial risk on returns positively, while the opposite holds for \textit{extroversion} and \textit{neuroticism}.

Different to these approaches, we focus on the \gls{mbti} rather than the Big 5. We create the first supervised model to predict \glspl{ceo}' \gls{mbti} personality from text by collecting a new dataset of crowd-annotated \gls{mbti} profiles. This sets us apart from prior work using unsupervised approaches trained on out-of-domain corpora.

\section{Personality Prediction}

Using transcribed speech data as an input, we predict the \gls{mbti} personality of \glspl{ceo} via text regression. The following sheds light on the dataset collection and validation, methodology, and results.

\subsection{Dataset Curation}
\label{data:pers_pred}
For this task, we collect data from two sources: (1) text data and (2) crowd-sourced personality data.

\paragraph{Text Data} We obtain 88K earnings call transcripts spanning years 2002--2020 from \textsc{Refinitiv Eikon}.\footnote{\url{https://eikon.thomsonreuters.com/index.html}} Earnings calls are quarterly teleconferences consisting of a scripted presentation and a spontaneous \gls{qa} session, in which \glspl{ceo} such as Elon Musk answer open questions of banking analysts. Due to the improvised nature of these answers, earnings calls are particularly suitable for detecting personal style \citep{doi:10.1177/0001839217712240}. Figure \ref{fig:earnings_call} shows an excerpt of Tesla's Q1 earnings call in 2020.

\begin{figure}[!t]
\begin{dialogue}
\footnotesize
\speak{Elon Musk (CEO)} Thank you. So Q1 ended up being a strong quarter despite many challenges in the final few weeks. This is the first time we have achieved positive GAAP net income in a seasonally weak first quarter. Even with all the challenges, we achieved a 20\% automotive gross margin, excluding regulatory credits, while ramping 2 major products. What we've learned from this is that\textemdash we've obviously learned a lot here.
\end{dialogue}
    \caption{Excerpt of Tesla's Q1 2020 earnings call.}
    \label{fig:earnings_call}
\end{figure}

Given the dialogue nature of the calls, we need to map utterances to individual \glspl{ceo} as we are not interested in the personality of the analysts. We identify \gls{ceo} names with regular expressions and minimal preprocessing (e.g., stripping middle name initials or titles). Next, we require a match with the executive database \textsc{Compustat Execucomp} for age and gender data (\S\ref{met:risk}),\footnote{\url{https://wrds-www.wharton.upenn.edu}} reducing our initial sample to 22K calls and 1.7K \glspl{ceo}. For these, we retrieve all of their utterances in the presentation and the \gls{qa} session of the calls.

\paragraph{Personality Data}
\label{data:pers}
We obtain \gls{mbti} personality labels for the \glspl{ceo} from \textsc{Personality Database},\footnote{\url{https://www.personality-database.com/}} which provides crowd-sourced personality profiles for celebrities, managers, and other noteworthy people. While each profile features vote results for the four dimensions of the \gls{mbti}, a minority also contains results for the Big 5. We find that 32 \glspl{ceo} (e.g., Elon Musk and Steve Jobs) from our earnings call sample have at least three \gls{mbti} votes available. The minimum, maximum, and mean votes per \gls{ceo} are 3, 1.8K, and 140, respectively. These \glspl{ceo} participate in a total of 736 earnings calls. Table \ref{tab:stats} gives the descriptive statistics of the merged text--personality data, and Table \ref{tab:mbti} contains example \glspl{ceo} from our dataset across the \gls{mbti}.

\begin{table*}[!t]
\centering
\footnotesize
\begin{tabular}{ll}
\toprule
MBTI & CEO Examples\\
\midrule
Extraversion & Steve Jobs (Apple), Lisa Su (AMD), Mary Barra (General Motors) \\
Introversion & Rupert Murdoch (Fox), Mark Zuckerberg (Facebook), Sheldon Adelson (Las Vegas Sands) \\
\midrule
Sensing & Jack Dorsey (Twitter), John Schnatter (Papa John's), Marcus Lemonis (Camping World)\\
Intuition & Marissa Mayer (Yahoo), Bob Iger (Disney), Evan Spiegel (Snap) \\
\midrule
Thinking & Elon Musk (Tesla), Tim Cook (Apple), Steve Ballmer (Microsoft)\\
Feeling & Sundar Pichai (Google), Howard Schultz (Starbucks), Naveen Jain (Infospace) \\
\midrule
Judging & Jeff Bezos (Amazon), Larry Ellison (Oracle), Martha Stewart (Martha Stewart Living)\\ 
Perceiving & Larry Page (Alphabet), Martin Shkreli (Retrophin), Donald Trump (Trump Entertainment)\\
\bottomrule
\end{tabular}
\caption{CEO examples for each MBTI dimension from our dataset.}
\label{tab:mbti}
\end{table*}

\begin{table}[!t]
\centering
\footnotesize
\begin{adjustbox}{max width=\linewidth}
\begin{tabular}{lS[table-format=7.0, round-mode=off]S[table-format=4.2, round-mode=places, round-precision=2]S[table-format=2.0, round-mode=off]S[table-format=4.0, round-mode=off]}
\toprule
Unit & $\Sigma_{x}$ & $\bar{x}$ & $\textrm{min}_{x}$ & $\textrm{max}_{x}$ \\
\midrule
utterances & 13183 & 17.911684782608695 & 2 & 124\\   
sentences & 111781 & 151.8763586956522 & 2 & 563 \\
tokens & 2526473 & 3432.7078804347825  & 22 & 9968\\
\bottomrule
\end{tabular}
\end{adjustbox}
\caption{Statistics of the \gls{ceo}--call data considered for the personality prediction. Sums ($\Sigma_{x}$), averages ($\bar{x}$), minima ($\textrm{min}_{x}$), and maxima ($\textrm{max}_{x}$) are computed across all earnings calls ($n = 736$).}
\label{tab:stats}
\end{table}

\begin{figure}[!t]
\centering
\includegraphics[width=\linewidth]{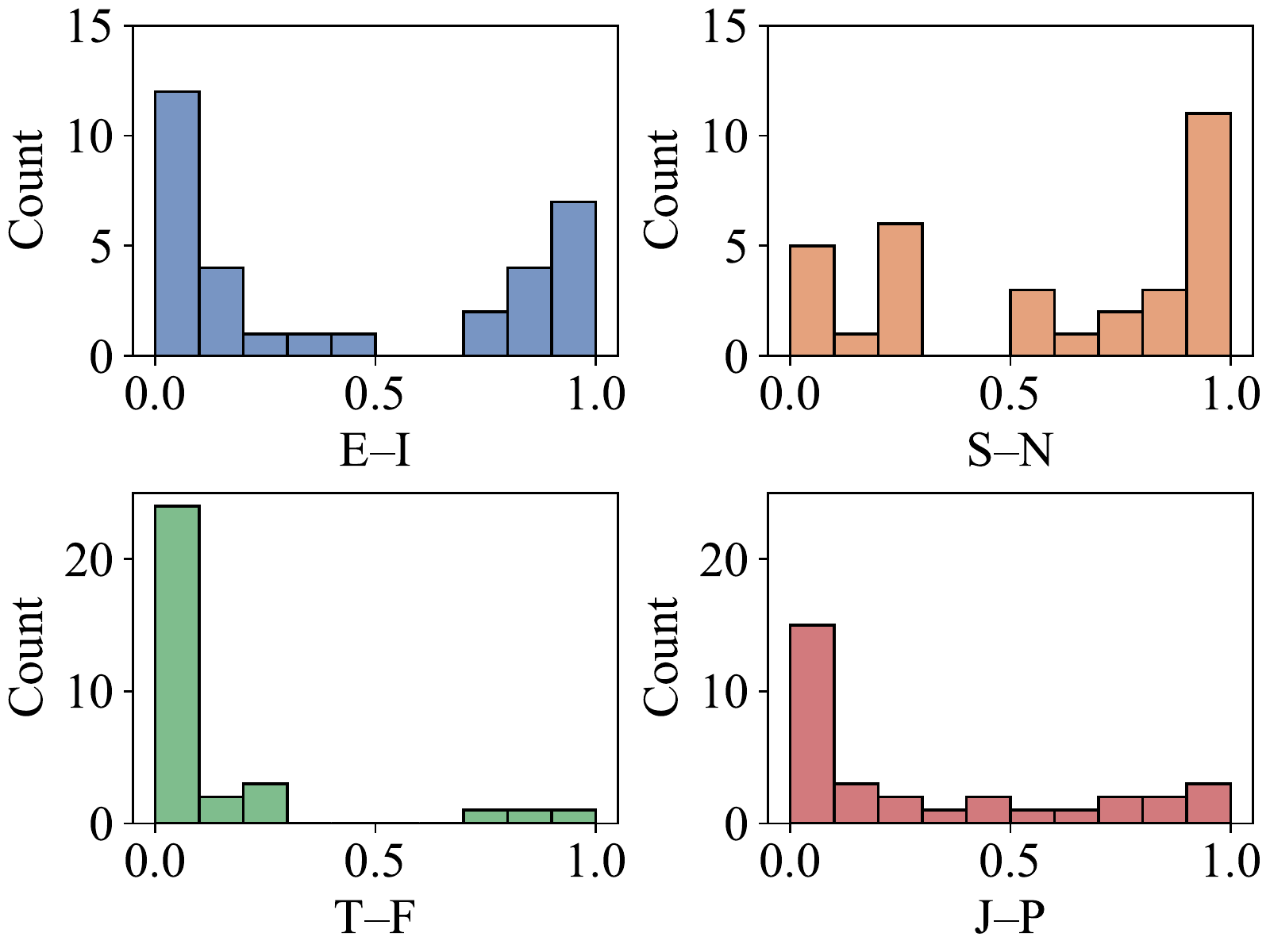}
\caption{Label distributions for all \glspl{ceo} considered in the personality prediction ($n = 32$) across the \gls{mbti} dimensions \textit{extraversion--introversion} (E--I), \textit{sensing--intuition} (S--N), \textit{thinking--feeling} (T--F), and \textit{judging--perceiving} (J--P).}
\label{fig:label_dist}
\end{figure}

Instead of representing each personality as one of 16 types, we represent each personality profile as a vector of 4 continuous variables ranging from 0 to 1, based on the crowd-sourced votes. We normalize the votes for the right-hand side of a scale $s$ by the total votes:
\begin{equation}
\mathrm{personality}_{s} = \frac{\mathrm{votes}_{1, s}}{\mathrm{votes}_{0, s} + \mathrm{votes}_{1, s}}.
\end{equation}
For example, for the E--I scale, we divide the votes for introversion (I) by the total votes for E and I. The resulting number is thus the likelihood of the \gls{ceo} being intro- or extroverted. This representation is similar to the Big 5 model (excluding the \textit{neuroticism} dimension) and allows for a more granular representation of personality than the usual operationalization of the \gls{mbti}. Figure \ref{fig:label_dist} shows the distributions of the such obtained continuous labels. Most CEOs in our sample are rather \textit{extroverted}, \textit{intuitive}, \textit{thinking}, and \textit{judging} (Figure \ref{fig:label_dist}), which corresponds to the ENTJ ``Decisive Strategist'' \gls{mbti} type.\footnote{\url{https://eu.themyersbriggs.com/en/tools/MBTI/MBTI-personality-Types/ENTJ}}

\paragraph{Internal Validation}
\label{met:iaa}
To assess the validity of the crowd-sourced votes, we analyze the inter-annotator agreement between the \gls{mbti} raters of the 32 \glspl{ceo} (Table \ref{tab:agreement}). While $p_{a}$ is high with values ranging between ca.\ 80 and 90\%, Krippendorff's $\alpha$ \cite{Krippendorff2013} yields only slight to moderate values between 0.14 and 0.43. \newcite{frequency_distribution} call this phenomenon the ``frequency distribution paradox,'' where highly skewed label distributions combined with high percentage agreements can lead to low values of $\alpha$. As measures robust to this undesirable property, they suggest the Brennan--Prediger coefficient $\kappa_{\textrm{bp}}$ \cite{brennan-prediger} and Gwet's $\gamma$ \cite{Gwet2008}, which in our case yield a high IAA between 0.60 to 0.88.

\begin{table}[!t]
\centering
\footnotesize
    \begin{tabular}{lS[round-mode=places, round-precision=2]S[round-mode=places, round-precision=2]S[round-mode=places, round-precision=2]S[round-mode=places, round-precision=2]}
\toprule

MBTI & $p_{a}$ & $\alpha$ & $\kappa_{\textrm{bp}}$ &  $\gamma$\\
\midrule
E--I & 87.45383 & 0.39865 & 0.74908 & 0.75633\\
S--N & 80.2042 & 0.42543 & 0.60408 & 0.6197 \\
T--F & 83.33429 & 0.13636 & 0.66669 & 0.7103 \\
J--P & 90.6237 & 0.17012 & 0.81247 & 0.88128\\
\bottomrule
    \end{tabular}
    \caption{\gls{iaa} per \gls{mbti} dimension in terms of percentage agreement ($p_a$), Krippendorff's $\alpha$, Brennan--Prediger coefficient ($\kappa_{\textrm{bp}}$), and Gwet's $\gamma$.}
    \label{tab:agreement}
\end{table}

\paragraph{External Validation}
\label{met:corr}

\begin{figure}
\centering
\includegraphics[width=\linewidth]{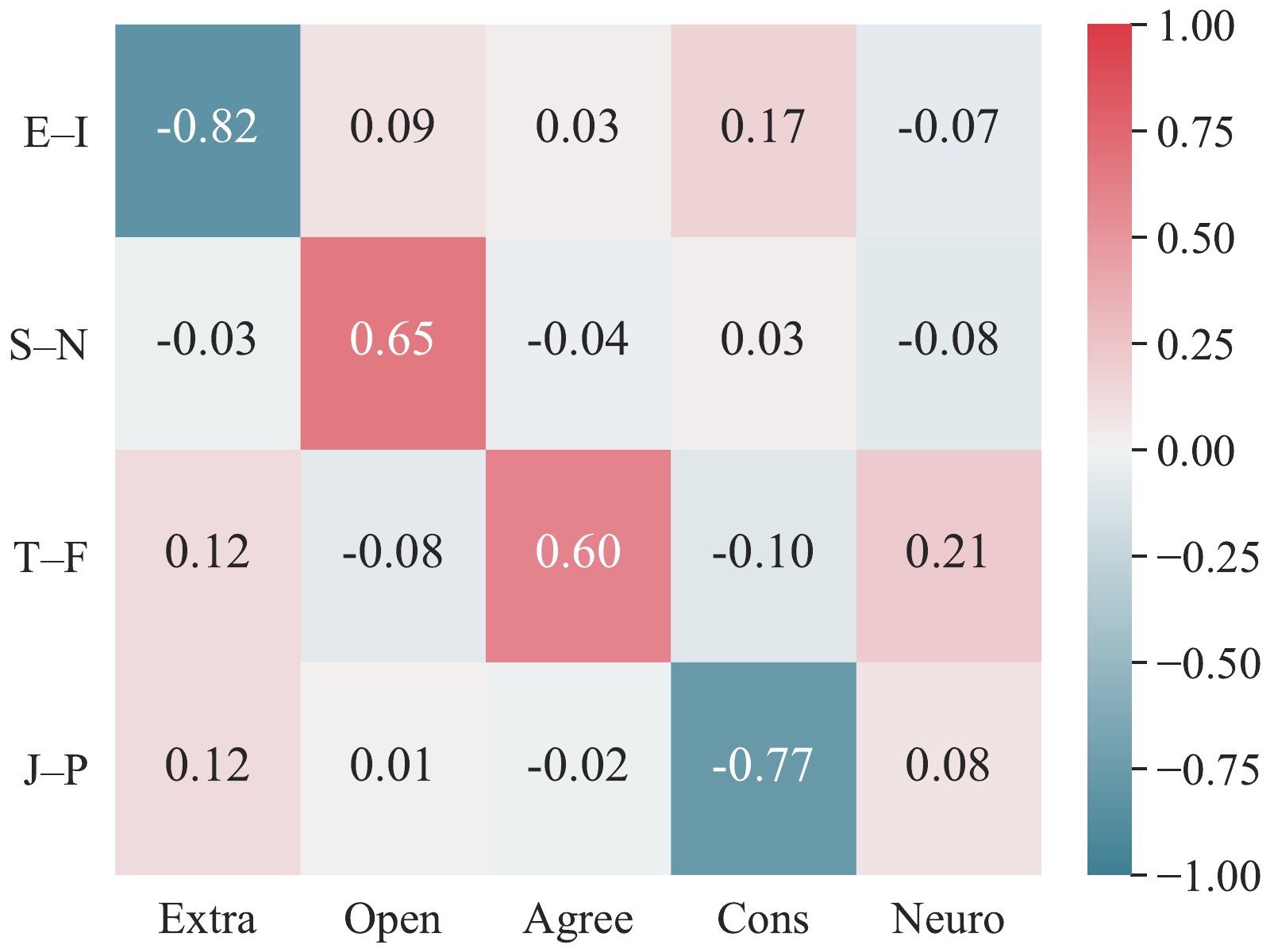}
\caption{Correlation of MBTI (y-axis) and Big 5 (x-axis) scales for all profiles on the \textsc{Personality Database} with at least three votes ($n = 2.2$K).}
\label{fig:corr}
\end{figure}

To get a notion of external validity, we construct a correlation matrix between the crowd-based \gls{mbti} and Big 5 votes of \textit{all} 2.2K profiles with more than three votes available on \textsc{Personality Database} (Figure \ref{fig:corr}). According to \newcite{McCrae1989} and subsequent work \cite{Furnham1996, Furnham2003}, strong correlations should exist between MBTI \textit{introversion} and Big 5 \textit{extraversion} ($r = -0.74$) as well as between MBTI \textit{intuition} and Big 5 \textit{openness} ($r = 0.72$). Furthermore, moderate correlations should exist between MBTI \textit{feeling} and Big 5 \textit{agreeableness} ($r = 0.44$) and between MBTI \textit{perceiving} and Big 5 \textit{conscientiousness} ($r = -0.49$). Our results confirm the findings of \newcite{McCrae1989} with similar correlations in the first two rows and stronger correlations in the third and fourth rows. This is most likely due to our increased sample size ($n = 2.2$K vs.\ $n=267$).

\subsection{Methodology}
\label{met:pers_pred}
For each of the 32 \glspl{ceo} appearing in 736 CEO--call instances, we compare sparse approaches suggested by past literature to Transformer architectures for a regression of \gls{mbti} personality.\footnote{The supplementary material contains our implementation and the earnings call identifiers. Using those, our corpus can be re-assembled from \textsc{Refinitiv Eikon}, \textsc{Seeking Alpha}, or alternative sources.}

\paragraph{Data Split} We apply an 80:10:10 split to our data to obtain separate training ($n = 568$), validation ($n = 84$), and test sets ($n = 84$). To avoid overfitting, we use \texttt{sklearn}'s \texttt{GroupShuffleSplit} with the \gls{ceo} names as group splitting criterion, i.e., we split the data such that no \gls{ceo} present in the training data appears in the validation or test data.

\paragraph{Normalization}
Given the highly skewed distributions, after the train--validation--test split, we apply a Box-Cox transformation \cite{Box1964} to $y$ with the following formula:
\begin{equation}
y(\lambda) = \begin{cases}
\frac{y^{\lambda} -1}{\lambda} & \text{for } \lambda \neq 0, \\ \ln(y) & \text{for } \lambda = 0.
\end{cases}
\end{equation}
We obtain $\lambda$ via maximum-likelihood estimation. The resulting transformation makes the four label distributions more Gaussian-like by stabilizing variance.

\paragraph{Transformers}
We explore cased-vocabulary BERT\textsubscript{base} (12-layer, 768-hidden, 12-heads, 109M parameters) \cite{Devlin2019} and RoBERTa\textsubscript{base} (12-layer, 768-hidden, 12-heads, 125M parameters) \cite{liu2019roberta} models with a linear regression head. The models are trained with a maximum sequence length of 512 and a sliding window approach. We determine the training batch size and learning rate by running a Bayesian optimization over the grid of batch sizes $b \in \{32, 64, 128, 256\}$ and learning rates $l \in [0, \num{5e-5}]$.\footnote{Final hyperparameter choices and results on our validation set can be found in Appendices \ref{app:params} and \ref{app:valid}.} We train a model for up to 10 epochs and early stopping with a patience of one epoch. For each of the four \gls{mbti} dimensions, we evaluate 40 combinations of hyperparameters and select the model with minimal loss on the validation set. Different to the \gls{mse} loss, which is implemented per default in the \huggingface Transformers \cite{wolf-etal-2020-transformers} regressors, we minimize the L1 or alternatively called \gls{mae} loss, which is less sensitive to outliers.

\paragraph{Sparse Methods} We also explore the sparse representations suggested by \newcite{Plank2015} and \newcite{Gjurkovic2018}. These include \gls{tfidf} vectors with $n$-grams of length $n \in \{1,2,3\}$ and dictionary features across all dimensions of \gls{liwc} 2015 \cite{liwc2015} fed into \gls{svm} and three-layer \gls{mlp} regressors. We compare all possible feature--algorithm combinations with respect to their average \gls{mae} on the validation set and select the combination with the lowest error (\gls{svm} with trigram \gls{tfidf}).

\paragraph{Evaluation} The final model performance is evaluated by inspecting the correlation and error between test set ground truth and prediction. As measures, we explore the linear correlation coefficient (i.e., Pearson's $r$) and the rank correlation coefficients Spearman's $\rho$ and Kendall's $\tau$. Instead of linear relationships, the latter two measure monotonic relationships and are more robust to outliers. In addition, we consider the error measure \gls{mae}, which is the minimized loss function of the Transformers. In case of a tie, we give precedence to $\tau$, as this measure is least sensitive to outliers and particularly suited for small sample sizes.

\subsection{Results and Discussion}
\label{res:pers_pred}
The results of the personality prediction task are depicted in Table \ref{tab:personality_prediction}. An \gls{svm} performs competitive, especially for the dimensions E--I ($\tau = 0.44$) and S--N ($\tau = 0.20$). While the \gls{svm} outperforms BERT for all dimensions except for J--P, RoBERTa achieves the best results in most cases.

The largest correlations across all models are achieved for the \textit{extraversion--introversion} (E--I) scale with strong linear and rank correlations for the RoBERTa regressor ($r = 0.70$, $\rho = 0.66$). This result is not surprising, as distinguishing between \textit{extra-} and \textit{introverted} \glspl{ceo} based on linguistic style should be comparably easy. This is followed by the \textit{sensing--intuition} (S--N) scale with moderate to strong correlations ($r = 0.45$, $\rho = 0.53$) and the \textit{judging--perceiving} (J--P) scale with weak to moderate correlations ($r = 0.40$, $\rho = 0.36$). The worst results are obtained for the \textit{thinking--feeling} (T--F) scale, with the \gls{svm} and RoBERTa obtaining correlations of around zero and BERT even obtaining weak to moderate negative correlations. There are several possible explanations for this: Conceptually, it could be the case that this dimension simply can not be captured by analyzing linguistic data. Furthermore, the predictive power could be low due to the comparably small sample size. Lastly, we hypothesize that the skewness of the label distribution, which was the highest across all \gls{mbti} dimensions for the T--F scale (Figure \ref{fig:label_dist}), has contributed to the weak performance. This warrants further research exploring whether our findings hold for larger datasets with less skewed label distributions.

\newcite{Stajner2020} hypothesize that the S--N and J--P dimensions should theoretically make for the worst candidates in a text-based personality prediction task since they capture behavioral rather than linguistic dimensions of personality. Although our regressors perform worse on these dimensions than for the \textit{extraversion--introversion} dimension, they still achieve moderate to strong correlations, showing that even the more latent dimensions of personality can be predicted from text.

\begin{table}[!t]
\centering
\begin{adjustbox}{max width=\linewidth}
\begin{tabular}{llS[table-format=0.2, round-mode=places, round-precision=2]S[table-format=0.2, round-mode=places, round-precision=2]S[table-format=0.2, round-mode=places, round-precision=2]S[table-format=0.2, round-mode=places, round-precision=2]S[table-format=0.2, round-mode=places, round-precision=2]S[table-format=0.2, round-mode=places, round-precision=2]}
\toprule
{MBTI}    & {Model} & {$r$}        & {$\rho$}      & {$\tau$}      & {MAE} \\
\midrule
                & {SVM}   & 0.568662426675648 & 0.5763365629649215 & 0.44230523605421745 & 0.38489306550492086 \\   
{E--I}   & {BERT} & 0.39332014599462595 & 0.3530214183872957 & 0.22322244094600896 & 0.5916627037556272\\\
                & {RoBERTa} & \bfseries 0.7024211144120937 & \bfseries 0.6560886362838613 & \bfseries 0.5184854047554954 & \bfseries 0.3399837404126423\\
\midrule
                & {SVM} & 0.3167056509864913 & 0.36126396939739447 & 0.19774769732877542 &  0.2993978962939358 \\
{S--N}   & {BERT} & 0.07721196397108031 & 0.2252101029128621 & 0.15826184472206342 & 0.4603169380373541\\
                & {RoBERTa}  & \bfseries 0.4454403260717685 & \bfseries 0.5294907077266214 & \bfseries 0.3779815084207027 & \bfseries 0.2764512316645347\\
\midrule
                & {SVM} & \bfseries 0.026398736793045528 & -0.11797500914450072 & -0.07785942403691712 & \bfseries 0.370752470172705\\
{T--F}   & {BERT} & -0.4681990990588438 & -0.4065905762455734 & -0.2746254998456951 & 0.4110043965575551 \\
                & {RoBERTa} &   0.00661499083662595 & \bfseries -0.09765070508904866 & \bfseries -0.06808627457621623 & 0.3859379094323412\\
\midrule
                & {SVM} & -0.046005799451036056 & 0.040052999129474944 & 0.02388257214115645 & \bfseries 0.3541108767453241\\
{J--P}   & {BERT} & 0.3851371830298552 & \bfseries  0.3811454994720046 & \bfseries 0.250607790334535 & 0.5217501833637788\\
                & {RoBERTa} & \bfseries 0.4015906232987949 & 0.3637491606567208 & 0.2079375947756688 & 0.364233414180257\\
\bottomrule
\end{tabular}
\end{adjustbox}
\caption{Correlation results of the personality regression task. \gls{ceo} personality is predicted across the \gls{mbti} dimensions \textit{extraversion--introversion} (E--I), \textit{sensing--intuition} (S--I), \textit{thinking--feeling} (T--F), and \textit{judging--perceiving} (J--P). \gls{svm} is trained on trigram \gls{tfidf} vectors, BERT\textsubscript{base}, and RoBERTa\textsubscript{base} on text. Best results in bold.}
\label{tab:personality_prediction}
\end{table}

\begin{figure*}[!t]
        \centering
        \begin{subfigure}[b]{0.475\textwidth}
            \centering
            \includegraphics[width=0.9\textwidth]{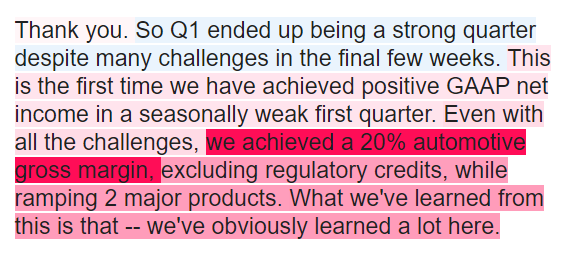}
            \caption{Result of the E--I regressor.}  
            \label{fig:shap_examples_ei}
        \end{subfigure}
        \hfill
        \begin{subfigure}[b]{0.475\textwidth}  
            \centering 
            \includegraphics[width=0.9\textwidth]{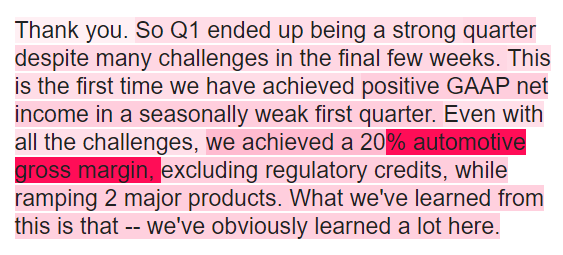}
            \caption{Result of the S--N regressor.} 
            \label{fig:shap_examples_sn}
        \end{subfigure}
        \vskip\baselineskip
        \begin{subfigure}[b]{0.475\textwidth}   
            \centering 
            \includegraphics[width=0.9\textwidth]{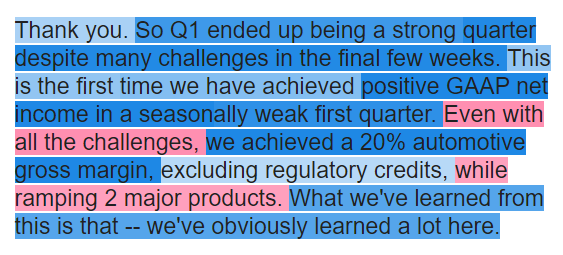}
            \caption{Result of the T--F regressor.}
            \label{fig:shap_examples_tf}
        \end{subfigure}
        \hfill
        \begin{subfigure}[b]{0.475\textwidth}   
            \centering 
            \includegraphics[width=0.9\textwidth]{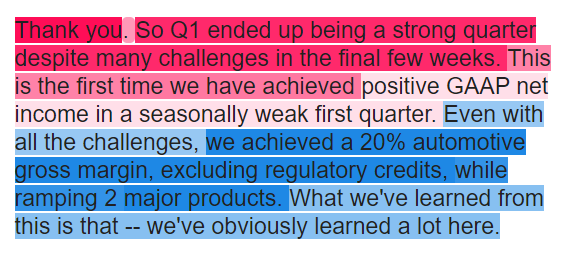}
            \caption{Result of the J--P regressor.}  
            \label{fig:shap_examples_jp}
        \end{subfigure}
\caption{Example snippet from our dataset (uttered by Elon Musk in Tesla's Q1 2020 earnings call) with SHAP heatmap across the \gls{mbti}. Red indicates a positive and blue a negative influence on the prediction.} 
\label{fig:shap_examples}
    \end{figure*}

\paragraph{Qualitative Analysis} As a brief qualitative analysis, we use Shapley Additive Explanations (SHAP) developed by \citet{NIPS2017_8a20a862} to visualize the personality predictions for an exemplary text snippet across the four \gls{mbti} dimensions with heatmaps (Figure \ref{fig:shap_examples}). The analyzed personality is Elon Musk, who, according to the crowd votes, scores high on E--I (\textit{introversion}) and on S--N (\textit{intuitive}), low on T--F (\textit{thinking}), and medium on J--P (\textit{judging/perceiving}). Particularly interesting are the results for T--F (Figure \ref{fig:shap_examples_tf}), where statements related to factual content are related to increased T, and interpretative statements (e.g., ``[e]ven with all the challenges'') to increased F.

\section{Risk Regression}
\label{risk_regression}
According to \textit{upper echelons theory} \cite{Hambrick1984}, strategic choices and performance measures of organizations can be predicted by characteristics of their top management. As a use case for our personality prediction task, we explore whether we can find empirical support for this theory. We hypothesize that having a different personality to most \glspl{ceo} (i.e., ENTJ, see Figure \ref{fig:label_dist} and \newcite{Cohen2013}) should translate into increased financial risk.

\subsection{Dataset Curation}
\label{data:fin}
As a basis for the risk regression task, we take the sample of 22K earnings calls and merge it with data obtained from the databases \textsc{CRSP}, \textsc{IBES}, and \textsc{Compustat Execucomp}, which we access via \textsc{WRDS}.\footnote{\url{https://wrds-www.wharton.upenn.edu}} To measure risk, we calculate the stock return volatility in the business week following each call as a label. We use the sample standard deviation of logarithmic stock returns for more robust measures. As features, we incorporate a comprehensive set of risk proxies suggested by \newcite{price12} and \newcite{Theil2019}.\footnote{We initially also considered including a market volatility index (VIX), but decided against it as its low explanatory power and high variation inflation factor (VIF) indicated redundancy of this variable \cite{vif}.} Furthermore, we include \gls{ceo} age and gender to control for possible confounding effects (e.g., being introverted could have a different effect for male than for female \glspl{ceo}). Definitions of all used controls are given in Table \ref{tab:finance}.

\begin{table}[!t]
\centering
\footnotesize
\begin{tabularx}{\linewidth}{p{0.15\linewidth}p{0.75\linewidth}}
\toprule
Feature & Definition \\
\midrule
Age & CEO age on the call date\\
Gender & CEO gender\\
Past Vola & Standard deviation of logarithmic returns in the business quarter before the call\\
Size & Market value of the firm, i.e., the number of outstanding shares times stock price one day before the call\\
Volume & Stock trading volume on the call date\\
Leverage & Total liabilities divided by assets\\
Spread & Difference between the stock's bid and ask price on the call date\\
BTM & Book-to-Market = book value of the firm divided by market value\\
SUE & Mean absolute deviation of analysts' earnings-per-share forecasts from the actual value in the preceding quarter\\
ROA & Return on Assets, i.e., net income divided by assets\\
Industry & Fama--French 12 industry dummies\\
Time & Year--quarter dummies\\
\bottomrule
\end{tabularx}
\caption{Controls used in the risk regression task. BTM is calculated following \cite{Fama2001} and firms with a negative value are removed. Size, BTM, and volume are log1p-transformed.}
\label{tab:finance}
\end{table}

\subsection{Methodology}
\label{met:risk}
We use the best-performing personality prediction model (RoBERTa) to infer the personality of the 1.7K unlabelled \glspl{ceo} present in the 22K calls. Together with the financial covariates (see above), the predicted \gls{ceo} \gls{mbti} is then used to explain short-term stock return volatility following the calls with multiple linear regression.\footnote{The supplementary material contains our dataset and implementation.} Volatility is the most common financial risk measure, and its prediction is an essential task for firm valuation and financial decision-making. Importantly, ``risk'' is a purely descriptive concept in finance, as it measures the fluctuation of stock returns.

\subsection{Results and Discussion}
\label{res:risk}
The results of this risk regression task are shown in Table \ref{tab:riskreg}. We find that the first three MBTI dimensions are significantly associated with risk following the call. This significance is high ($p \leq 0.001$) for E--I and T--F. The direction of this association behaves as expected: a \gls{ceo} communicating in an \textit{introverted} and \textit{feeling} manner is associated with increased risk ($\beta_{i} = 0.03$, $\beta_{f} = 0.10$, while an \textit{intuitive} communication is associated with decreased risk ($\beta_{s} = -0.02$). Notably, these results are robust to age- and gender-fixed effects. Although seemingly small, the size of the personality effect (i.e., the coefficient height) is in line with that observed by related work \citep{doi:10.5465/amj.2018.0626}. It is expectable that fundamentals such as past risk or firm size have a stronger impact on future risk than, e.g., \gls{ceo} extraversion. Remarkably, T--F has the third-largest impact ($\beta_{f} = 0.10$) out of all considered features. Though only weakly correlated with the ground truth (Table \ref{tab:personality_prediction}), the results suggest that the predictions for this scale contain strong economic signal for risk regression.

\sisetup{
    table-format=2.3,
    round-integer-to-decimal = true,
    group-digits             = true,
    group-minimum-digits     = 4,
    group-separator          = {\,},
    table-align-text-pre     = false,
    table-align-text-post    = false,
    input-signs              = + -,
    input-symbols            = {*} {**} {***},
    input-open-uncertainty   = ,
    input-close-uncertainty  = ,
    retain-explicit-plus
}
\begin{table}[!t]
\centering
\footnotesize
\begin{tabular}{lS[table-format=2.2, round-mode=places, round-precision=2, table-space-text-pre={**}, table-space-text-post={-**}]S[table-format=2.2, round-mode=places, round-precision=2,table-space-text-pre={**}, table-space-text-post={-**}]}
\toprule
Feature & {\textsc{Fin}} & {\textsc{Fin} + \textsc{MBTI}} \\
\midrule
E--I     &         & 0.0317$^{***}$\\
                 &         & (5.007)   \\
S--N      &         & -0.0168$^{**}$ \\
                 &         & (-2.688)  \\
T--F      &         & 0.1010$^{***}$ \\
                 &         & (13.673)  \\
J--P      &         & -0.0016 \\
                 &         & (-0.220)  \\
\addlinespace[4pt]
Age & & -0.0052\\
& & (0.377) \\
Gender             &  &  -0.0185 \\
& & (-0.748) \\
\addlinespace[4pt]
Past Vola & 0.4352$^{***}$& 0.4257$^{***}$  \\
                 & (45.801)  & (44.724)   \\
Size             & -0.1840$^{***}$ & -0.1920$^{***}$ \\
                 & (-19.065) & (-19.826)  \\
Volume              & 0.0445$^{***}$  & 0.0450$^{***}$  \\
                 & (5.282)   & (5.360)   \\
Leverage              & -0.0573$^{***}$  & -0.0460$^{***}$  \\
                 & (-8.675)   & (-6.883)   \\
Spread              & 0.0271$^{***}$  & 0.0257$^{***}$  \\
                 & (4.304)   & (4.097)   \\   
BTM             & -0.0421$^{***}$  & -0.0207$^{***}$  \\
                 & (-6.220)   & (-2.916)   \\                 
SUE             & -0.0023  & -0.0041  \\
                 & (-0.411)   & (-0.732)   \\
ROA             & -0.0012  &  0.0027 \\
                 & (-0.207)   & (0.455)   \\                 
\addlinespace[4pt]
$n$ & {21,787} & {21,787} \\
Adj. $R^{2}$      & {33.40\%} & {34.00\%} \\
\bottomrule
\addlinespace[3pt]
\multicolumn{3}{c}{\footnotesize $^{*} p \leq 0.05$, $^{**} p \leq 0.01$, $^{***} p \leq 0.001$}\\
\end{tabular}
\caption{Results of the risk regression with $z$-standardized coefficients and $t$-statistics in parentheses. The sample consists of 22K earnings calls spanning 1.7K firms and years 2002--2020. Regressions include fixed effects for industry and time. \textsc{Fin} is a model with just the financial features (defined in \S\ref{data:fin}) and \textsc{Fin} + \textsc{MBTI} is a joint model including the MBTI (E--I, S--N, T--F, and J--P) along with CEO age and gender.}
\label{tab:riskreg}
\end{table}

In sum, these results provide new empirical evidence to support the \textit{upper echelons theory}. We show that situational aspects of \gls{ceo} personality, predicted with our MBTI regressor, also reflect firm performance measured by stock return volatility, the most common financial risk measure.

\section{Ethical Considerations}
\label{sec:ethics}
In the following, we discuss possible biases and environmental considerations.

\paragraph{Social Desirability Bias} Past literature has shown that some Big 5 personalities are more socially desirable than others, which paves the way to discrimination: Overall, it is socially desirable to score low on \textit{neuroticism} (an omitted scale in the \gls{mbti}) and high on \textit{conscientiousness} and \textit{agreeableness}. To a lesser extent, it is socially desirable to score high on \textit{extraversion} and \textit{openness} \cite[Table 2]{Ones1996}. For the MBTI, in contrast, there exist no ``bad'' personality traits. As shown in \S\ref{met:corr}, however, the Big 5 and the MBTI correlate. Therefore, the points raised about social desirability, albeit to a lesser extent, should apply here, too.

\paragraph{Sample Biases} Critically, our gold standard consists of just 32 \glspl{ceo} of large American (mostly tech) companies. While these companies (Alphabet, Facebook, Apple, etc.) constitute a large share of the American market, this renders the personality prediction model less applicable to non-American, small, or non-tech companies. Only four (i.e., 12.5\%) of the 32 \glspl{ceo} are female. While this gender ratio is twice as high as that of the S\&P 500 \cite{catalyst}, this highlights that the findings of this study might generalize poorly to non-male \glspl{ceo}. In addition, as shown in \S\ref{data:pers}, Figure \ref{fig:label_dist}, \glspl{ceo} as a social cohort share a distinct distribution of personality traits, which is why we argue that the \gls{mbti} regressors should only be applied with caution, if at all, to non-executive samples.

\paragraph{Energy Consumption}
Training neural models can have substantial financial and environmental costs \citep{Strubell2019}, which motivates us to discuss the computational efficiency of the Transformers. Using an NVIDIA Tesla P100 GPU, we run a hyperparameter optimization over 40 configurations  per \gls{mbti} dimension for both BERT and RoBERTa. The average power consumption is 200W and the optimization takes ca.\ 16 hours, i.e., 3.2 kilowatt hours (kWh) with an electricity cost of 40 cents per model.\footnote{Calculations assume the average U.S. electricity rate of 12.55 cents per 15 November 2021: \url{https://www.electricchoice.com/electricity-prices-by-state}} Labeling the 22K earnings call instances with no available ground truth takes ca.\ 4.5 hours and 140W, i.e., 0.63 kWH of GPU time and 8 cents, respectively. Training time of the \gls{svm} with trigram \gls{tfidf} is negligible (ca.\ 2 minutes on a quad-core processor with 8GB RAM). Whether the performance increases of the Transformers over a sparse method justify the added computational costs should be considered carefully on a case-by-case basis.

\section{Conclusion and Future Work}
We present the first text regression approach for predicting the \gls{mbti} personality of \glspl{ceo}. Although past research has contested the possibility of predicting \gls{mbti} from purely textual data, we observe moderate to strong correlations with the ground truth for three out of four dimensions. In a risk regression task, we demonstrate that\textemdash consistent with the \textit{upper echelons theory}\textemdash the predicted \gls{ceo} personality is significantly associated with financial risk in the form of stock return volatility. Qualitatively, extroverted, intuitive, and thinking \glspl{ceo} seem to incur less financial risk.

In the future, we plan to  model the personality prediction task as a multi-task learning problem, in which one single regressor is trained to predict all four MBTI dimensions at once. In addition, it would be interesting to incorporate speech signals of executives (e.g., voice modulation, tonality, and silence) into the personality predictions.

\section*{Acknowledgments} We would like to thank Amanda Cercas Curry, Federico Bianchi, Tommaso Fornaciari, and Anne Lauscher for their helpful feedback on an earlier version of this paper. Furthermore, we are grateful to all other members of MilaNLP Lab at Bocconi University for the fruitful discussions.

\bibliography{references}
\bibliographystyle{acl_natbib}

\clearpage

\appendix

\section{Hyperparameter Configurations}
\label{app:params}
Using a Bayesian hyperparameter optimization as specified in \S\ref{met:pers_pred}, the following configurations led to minimal loss on the validation set. Table \ref{tab:params_bert} summarizes the optimal configuration for BERT and Table \ref{tab:params_roberta} the one for RoBERTa.

\begin{table}[H]
\begin{subtable}{1\linewidth}
    \centering
    \begin{tabular}{lS[table-format=3]S[round-mode=figures,scientific-notation=true,table-format=1.1e-1]}
    \toprule
    {MBTI} & {Batch Size} & {Learning Rate}\\
    \midrule
      E--I & 128 & 0.000047709367240367057\\
      S--N  & 32 & 0.00004870298833658933\\
      T--F   & 32 & 0.0000010197034363264246\\
      J--P   & 256 & 0.000008609163790147124\\
     \bottomrule
    \end{tabular}
    \caption{Hyperparameters for BERT.}
    \label{tab:params_bert}
\end{subtable}
\par\bigskip
\begin{subtable}{1\linewidth}
    \centering
    \begin{tabular}{lS[table-format=3]S[round-mode=figures,scientific-notation=true,table-format=1.1e-1]}
    \toprule
    {MBTI} & {Batch Size} & {Learning Rate}\\
    \midrule
      E--I   & 256 & 0.00004291971286455248\\
    S--N     & 32 & 0.00004580768232115495\\
    T--F & 128 & 9.399749171372097e-8\\
    J--P & 128 & 0.0000467260784279696 \\
    
    \bottomrule
    \end{tabular}
    \caption{Hyperparameters for RoBERTa.}
    \label{tab:params_roberta}
\end{subtable}
\caption{Final hyperparameter configurations found by the Bayesian optimization searching over 40 configurations per \gls{mbti} dimension.}
\end{table}

\section{Results on the Validation Set}
\label{app:valid}
The results of the \gls{mbti} regressors on the validation set are depicted in Table \ref{tab:valid_results}.

\begin{table}[H]
\centering
\begin{adjustbox}{max width=\linewidth}
\begin{tabular}{llS[table-format=0.2, round-mode=places, round-precision=2]S[table-format=0.2, round-mode=places, round-precision=2]S[table-format=0.2, round-mode=places, round-precision=2]S[table-format=0.2, round-mode=places, round-precision=2]S[table-format=0.2, round-mode=places, round-precision=2]S[table-format=0.2, round-mode=places, round-precision=2]}
\toprule
{MBTI}    & {Model} & {$r$}        & {$\rho$}      & {$\tau$}      & {MAE} \\
\midrule
                & {SVM}   & 0.7018983376476876 & 0.6896822456960616 & 0.5483603115634476 & 0.3765243771711043 \\   
{E--I}   & {BERT} & 0.45619431670956945 & 0.41742255807883666 & 0.2812599250624726 & 0.6227526825382321\\
                & {RoBERTa} & 0.7184024822236219 & 0.5999344195720201 & 0.48206792648007313 & 0.3507149192386135\\
\midrule
                & {SVM} & 0.34197221535449607 & 0.48201103881012625 & 0.2992811559589239 & 0.28203906332373485 \\
{S--N}   & {BERT} & 0.20399213726155072 & 0.34850737718905817 & 0.242643001712934 & 0.5341430028715195\\
                & {RoBERTa}  & 0.42916016598997386 & 0.6056444156946413 & 0.43057869534735504 & 0.26503335530888344\\
\midrule
                & {SVM} & 0.13126972504584072 & -0.04531403862321805 & -0.029090699904967657 & 0.33293868429750056\\
{T--F}   & {BERT} &-0.4298315348327885 & -0.31679060081089877 & -0.21932426557565504 & 0.38233224390252357 \\
                & {RoBERTa} &   0.11063016597481194 & -0.06771495783188156 & -0.03301304146518802 & 0.3593431958747135\\
\midrule
                & {SVM} & -0.045894291494788124 & 0.049330622239695074 & 0.030893538679630853 & 0.34558121342270975\\
{J--P}   & {BERT} & 0.32129240115377306 & 0.28059066556662315 & 0.18857930902358 & 0.525056858384407\\
                & {RoBERTa} & 0.24957817730107068 & 0.144778043764823 & 0.057281769635148874 & 0.39852700445353856\\
\bottomrule
\end{tabular}
\end{adjustbox}
\caption{Results of the personality prediction task on the validation set.}
\label{tab:valid_results}
\end{table}

\end{document}